\documentclass[letterpaper, 10 pt, conference]{ieeeconf}  

\usepackage{changes}

\usepackage{todonotes}
\usepackage{personal}

\usepackage{algorithmic,algorithm2e,float}

\IEEEoverridecommandlockouts
\newcommand{\cupdot}{\mathbin{\mathaccent\cdot\cup}}

\newcommand{\wtfargmax}[1]{\underset{#1}{\operatorname{arg}\,\operatorname{max}}\;}

\begin{document}
    \title{
            Object Placement Planning and Optimization for Robot Manipulators
     }
     \author{Joshua A. Haustein${}^{1}$, Kaiyu Hang${}^{2}$, Johannes Stork${}^{3}$ and Danica Kragic${}^{1}$
         \thanks{${}^{1}$ Division of Robotics, Perception and Learning (RPL), CAS, CSC,
                    KTH Royal Institute of Technology,
                    Stockholm, Sweden,
                    E-mail: haustein, dani@kth.se}
         \thanks{${}^{2}$GRAB Lab, Yale University, New Haven, USA, E-mail: kaiyu.hang@yale.edu}
         \thanks{${}^{3}$Center for Applied Autonomous Sensor Systems (AASS), Örebro University, Örebro, Sweden, E-mail: johannesandreas.stork@oru.se}
       }

    \maketitle
\begin{abstract}
We address the problem of motion planning for a robotic manipulator with the task to place
a grasped object in a cluttered environment.
In this task, we need to locate a collision-free pose for the object that \textit{a)} facilitates
 the stable placement of the object, \textit{b)} is reachable by the robot manipulator and \textit{c)} optimizes a user-given placement objective. Because of the placement objective, this problem is more challenging than classical motion planning where the target pose is defined from the start. To solve this task, we propose an anytime algorithm that integrates sampling-based motion planning for the robot manipulator with a novel hierarchical search for suitable placement poses. We evaluate our approach on a dual-arm robot for two different placement objectives, and observe its effectiveness even in challenging scenarios. 
\end{abstract}

\section{Introduction}
Pick-and-place is among the most common tasks robot manipulators are applied for today.
Grasp planning, which is the process of autonomously selecting grasps, still receives much attention and effort from the robotics community~\cite{bicchi2000robotic, Bohg14, Roa2015}.
In contrast, the problem of placement planning, which is the process of autonomously deciding where
and how to place an object with a robot, has received considerably less
attention.

An autonomous robot tasked with placing a grasped object can generally not assume to know the environment
in advance, rather it faces the following challenges when perceiving the environment for the first time:
\begin{itemize}
 \item[1.] It needs to identify suitable locations that afford placing.
        For instance, an object may be placed flat on a horizontal surface,
        leaned against a wall, placed on a hook, or laid on top of other objects.
        Determining how and where a particular object can be placed, requires analysis
        of both the environment's and the object's physical properties.
  \item[2.] It needs to be able to reach the placement.
          Placing requires the robot to move close to obstacles, which make it
          difficult to compute collision-free arm configurations reaching a placement. In addition,
          the obstacles render planning an approach motion computationally expensive.
  \item[3.] Not all placements are equally desirable. For many tasks, there exists an objective
          such as stability, human-preference on location or clearance from other obstacles, that is
          to be maximized.
\end{itemize}

\begin{figure}
        \setlength{\tabcolsep}{2pt}
        \begin{tabular}{cc}
                        \includegraphics[width=0.48\columnwidth]{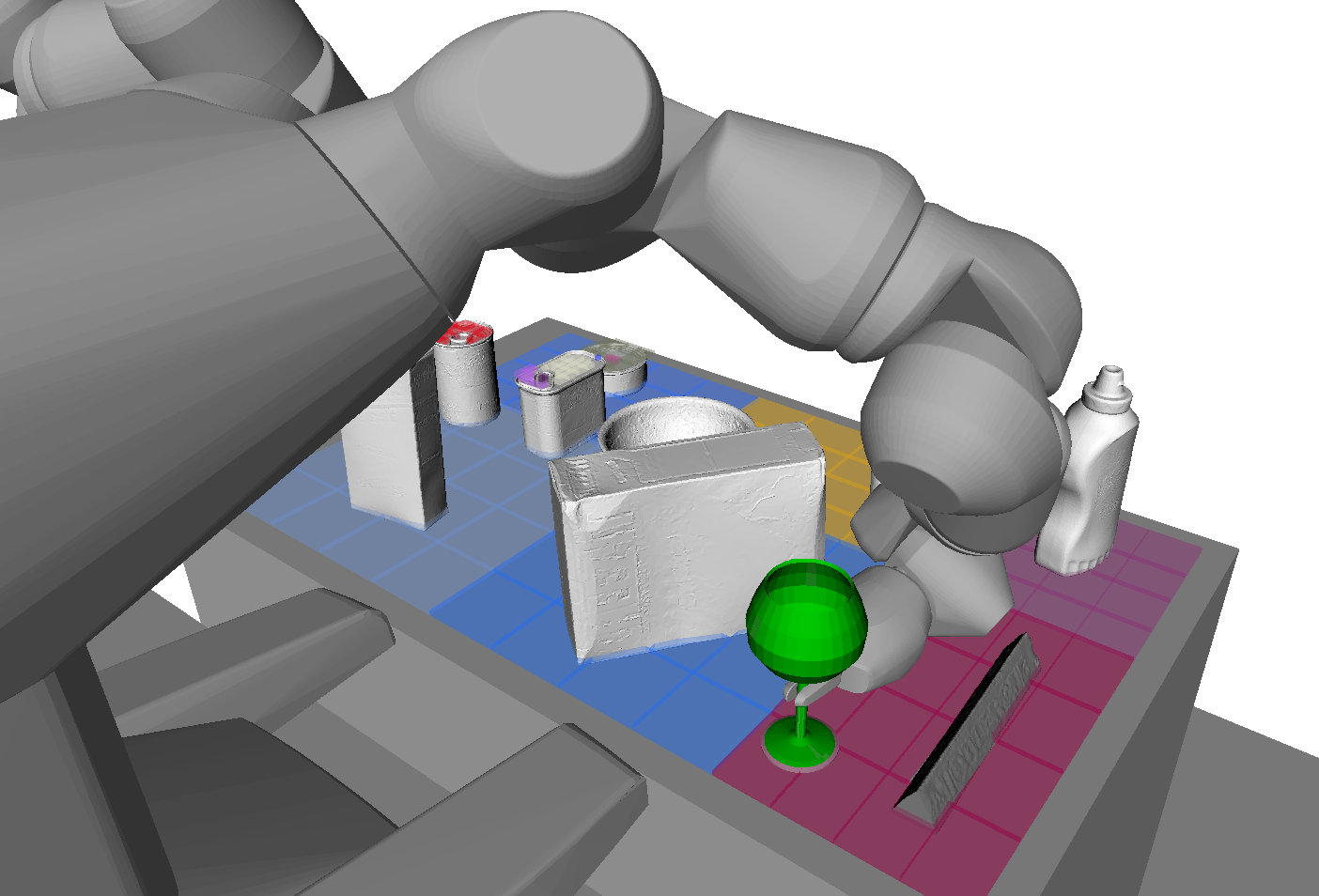}
                &
                        \includegraphics[width=0.48\columnwidth]{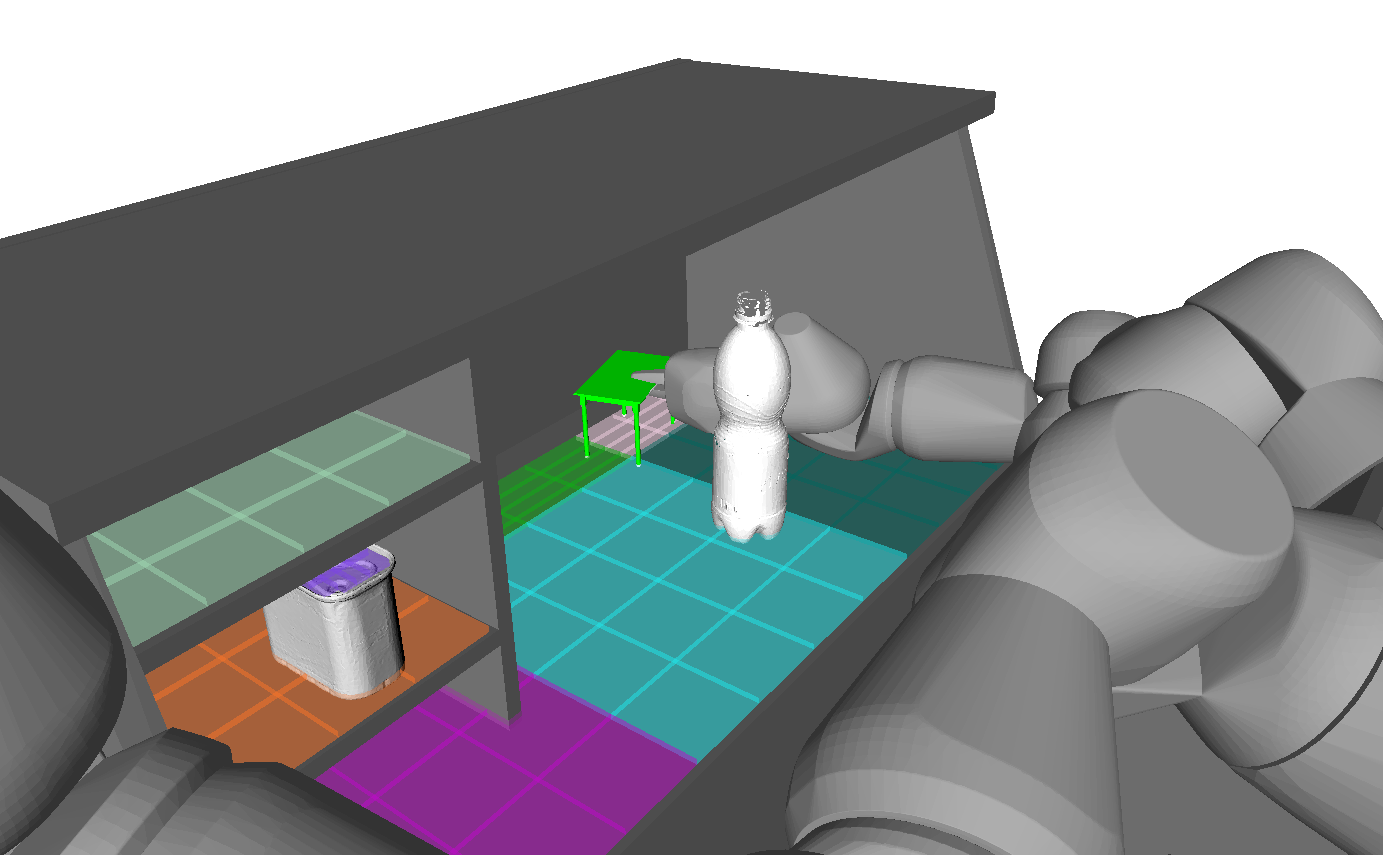}
                \\
                        \includegraphics[width=0.48\columnwidth, trim={0 0 0 0.4cm}, clip]{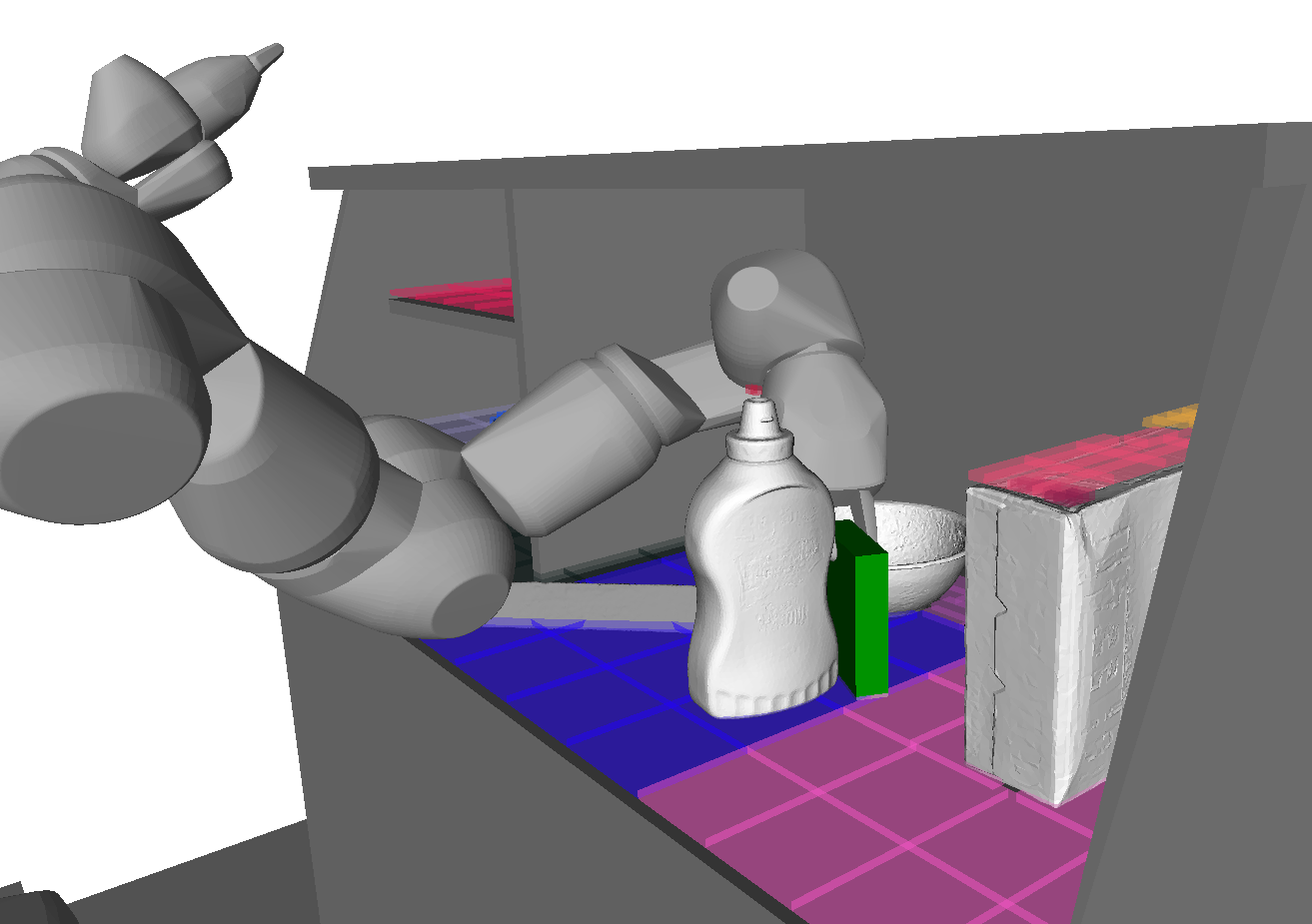}
                &
                        \includegraphics[width=0.48\columnwidth, trim={0 0 0 0.4cm}, clip]{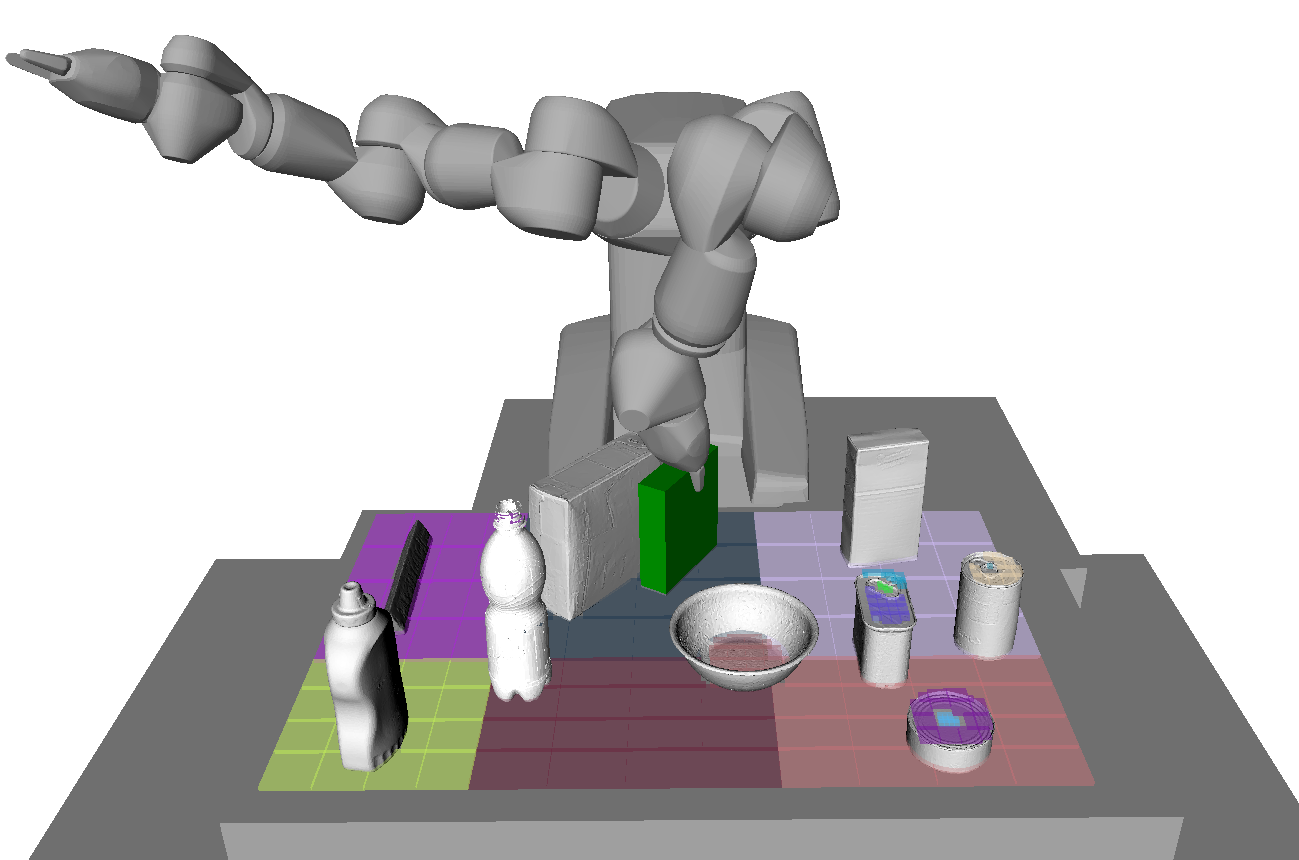}
        \end{tabular}
        \caption{Our algorithm computes placements for objects as well as corresponding approach motions
                in cluttered environments. In addition, it optimizes a user-specified objective for the placement
                pose. In the top row are example placements produced by our algorithm for a wine glass and toy table (\textit{green})
                under the objective to maximize clearance from other objects. In the bottom row
                a small and a large crayons box (\textit{green}) are placed under the objective to minimize clearance.}
        \label{fig:figure_one}
        \vspace{-0.5cm}
\end{figure}

Our contribution is an algorithmic framework that addresses these challenges and our main focus
lies on computing reachable placement poses (challenge 2) that maximize a user-specified objective
(challenge 3). In particular, we consider a dual-arm robot in difficult to navigate environments, such as
cluttered shelves and cupboards, \figref{fig:figure_one}. Our approach addresses challenge 2 by
integrating a motion planning algorithm with a novel hierarchical search for a placement pose.
We address challenge 3 by designing the algorithm such that it finds an initial feasible solution quickly,
and then incrementally improves the user-specified objective in an anytime fashion.

\section{Related Works} \label{sec:related_works}
\subsection{Placing objects}
Previous works on placing objects predominantly focus on challenge 1, i.e.\
searching poses in the environment, where an object can rest stably.
A naïve solution consists in identifying horizontal surfaces in the environment
and placing the object flat on the surface where there is enough space.
This technique is, for instance, commonly employed in manipulation planning works
which focus on planning complex sequences of pick-and-place
operations rather than individual placements~\cite{Schmitt2017, Xian2017, Garrett2018,
Wan2018, Lertkultanon2018}.

The object's orientation for a horizontal placement can be obtained
by analyzing the object's convex hull and extracting the faces that support a stable
placement~\cite{Wan2018, Lertkultanon2018}. Each of these faces gives
rise to a base orientation when aligned with the support surface.
Different poses with the same base orientation can then be
obtained by translating the object on the support surface and rotating it around the surface's
normal.  To locate collision-free and reachable placement poses
(challenges 1 and 2), these works then employ rejection sampling of positions and
orientations using a collision-checker and inverse kinematics solver.  This is
sufficiently efficient, if there are few obstacles
and most sampled poses are within reach.  If this is not the case, however, a more efficient
strategy such as the one presented in this work is required.

More complex approaches to locating placement poses (challenge 1) have been presented by
Schuster et al.\ \cite{Schuster2010}, Harada et al.\ \cite{Harada2014} and
Jiang et al.\ \cite{Jiang2012}. Schuster et al.\
present a data-driven segmentation algorithm to discriminate clutter from support surfaces,
and apply this segmentation to extract candidate placement poses.
Similarly, Jiang et al. follow a data-driven approach and train a classifier to score
the placement suitability of candidate poses based on manually defined features.
These features are extracted from 3D point-clouds of the object and the environment, and include physical
feasibility, stability, as well as human placement preference. The approach is capable
of identifying a variety of placements, such as placing a plate in a dish-rack, hanging a mug on a bar
or laying a box on a flat surface. In order to evaluate the classifier, however, the approach requires
a set of candidate poses. Obtaining these in cluttered environments is non-trivial, as
random sampling, for instance, has low probability of sampling good candidate.

Harada et al.\ \cite{Harada2014} locate placement poses by matching planar surface patches on the object with
planar surface patches in the environment. This allows the approach to locate placements on
large, flat surfaces and also placements where the handle of a mug is hanging on a flat bar.
While the work also integrates this algorithm with a motion planner (challenge 2),
it does not perform any optimization of an objective (challenge 3).

In contrast to these previous works, our work's focus lies on computing reachable
placement poses among obstacles (challenge 2) that maximize a user-specified objective (challenge 3).
We follow aforementioned previous works when addressing challenge 1 and place objects
on horizontal support surfaces.

\subsection{Integrated grasp and motion planning}
Ensuring that a collision-free approach motion to a placement exists (challenge 2)
requires us to closely integrate the placement search with a motion planning algorithm.
This relates our problem to integrated grasp and motion
planning~\cite{Haustein2017, Rosell2013, berenson2007grasp, Vahrenkamp2012, Fontanals2014}.
These works present algorithms that simultaneously compute grasps with corresponding approach motions,
and demonstrate that in cluttered environments separate
planning of grasps and approach motions is inefficient. This is due to the fact that
many potential grasps are in collision or out of reach. Our work addresses the analogous challenge
for placing, with the extension of optimizing an objective function on the placement (challenge 3).
The method that we employ relates to our previous work on integrated grasp and motion planning~\cite{Haustein2017}.

\section{Problem Definition}
\label{sec:problem}
We consider a dual-arm robot equipped with two manipulators, $\mathcal{A} = \{1, 2\}$, that
is tasked to place a \emph{rigid} object $o$ in a user-defined target volume $V \subset \mathbb{R}^3$
in its workspace. We assume that the object can be grasped by either arm, and the process
of acquiring a stable grasp is known a priori.
The target volume $V$ is a set of positions for the object $o$ and
restricts the search space for placement poses to
$\mathcal{X}^o = V \times \textit{SO}(3) \subset \textit{SE}(3)$.
Obviously, not all poses in $\mathcal{X}^o$
facilitate a stable placement, since for many of these the object might be, for example,
in midair or intersecting obstacles.
We denote the constraint that a pose $\bm{x} \in \mathcal{X}^o$ facilitates
the stable placement of the object as binary mapping $s(\bm{x})$, that is $1$ if $\bm{x}$ is a stable
placement and $0$ otherwise. Additionally, we denote the constraint that a pose $\bm{x}$
is physically feasible,
i.e.\ that there is no intersection of the interior of the object with any obstacle,
as binary predicate $c_f(\bm{x})$.

A placement pose must be reachable by the robot manipulator.
For this, let $\mathcal{C}^a = \mathcal{C}^a_\text{free} \cupdot \mathcal{C}^a_\text{obst}$ denote
the configuration space of arm $a \in \mathcal{A}$, and let
$O(q) \colon \mathcal{C}^a \to \textit{SE}(3)$ denote the pose $\bm{x} \in \textit{SE}(3)$ of the grasped
object when the arm is in configuration $q \in \mathcal{C}^a$.
We say a pose $\bm{x} \in \mathcal{X}^o$ is \textit{path-reachable}, $r(\bm{x}) = 1$, if
for some arm $a \in \mathcal{A}$ there exists a known collision-free continuous path
$\tau\colon [0, 1] \to \mathcal{C}^a_\text{free}$
starting from the initial configuration of the robot $\tau(0)~=~q_0~\in~\mathcal{C}^a_\text{free}$
and ending in a configuration $\tau(1) = q_g \in \mathcal{C}^a_\text{free}$ such that it reaches
$\bm{x}$, i.e.\ $O(q) = \bm{x}$.

With these constraints and a user-provided objective
function $\xi: \mathcal{X}^o \to \mathbb{R}$, we formalize our task as the following
constraint optimization problem:

\begin{equation}
        \label{eq:general_plcmnt}
\begin{aligned}
        & \underset{\bm{x} \in \mathcal{X}^o}{\text{maximize}} & &  \xi(\bm{x}) \\
        & \text{subject to} & &c_f(\bm{x}) &= 1 \\
        & & &s(\bm{x}) &= 1 \\
        & & &r(\bm{x}) &= 1
\end{aligned}
\end{equation}
Independently of the objective function, the optimization problem is challenging to solve due
to the constraints. The collision-free constraint $c_f(\bm{x})$ renders the problem non-convex.
The stability constraint, $s(\bm{x})$, is difficult to model, as it is a function of the
physical properties of the object and the local environment.
Lastly, the path-reachability constraint $r(\bm{x})$ requires a motion planning
algorithm to compute an approach path, which is generally computationally expensive.

\begin{figure*}[ht]
\includegraphics[width=\textwidth]{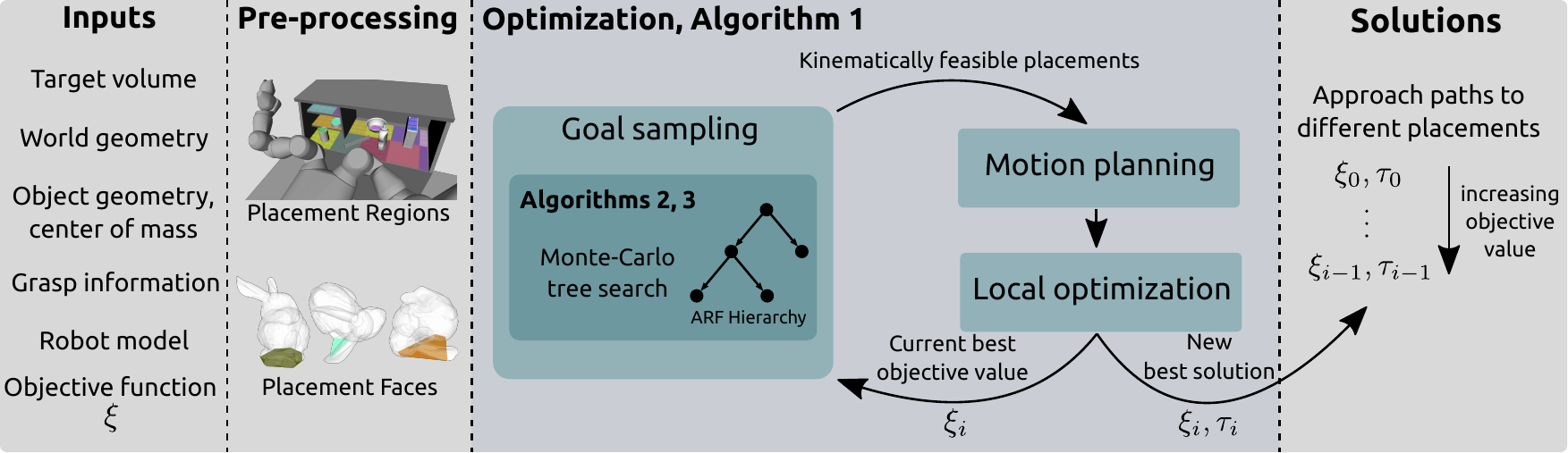}
        \caption{Our approach consists of two stages. In a pre-processing stage we first extract
                placement regions and faces that help locating us stable object poses. In the optimization
                stage a sampling algorithm is employed to locate kinematically reachable and collision-free
                stable placement poses. These are provided to a motion algorithm to verify path-reachability
                and construct an approach motion. Subsequently a local optimization algorithm is employed
                to improve the placement locally. Any found solution is made available to the user, and
                subsequent iterations search for better solutions.}
        \label{fig:overview}
        \vspace{-0.56cm}
\end{figure*}

Note that after releasing an object at a placement pose, the robot might not be able to retreat without
colliding with the placed object. Hence, in principle, there is the additional constraint that a
collision-free retreat motion must be possible. In this work, however, we choose to exclude this constraint
from our problem definition and instead assume that a collision-free retreat is always possible.

\subsection{Assumptions on prior information}
We assume access to the kinematic and
geometric model of the robot, the geometry of the object, the location of its center of mass,
and the geometry of the environment in form of surface points $\mathcal{S} \subsetneq \mathbb{R}^3$.
Furthermore, we assume that the environment is rigid, and that gravity acts antiparallel to the
$z$-axis of the world's reference frame.

We also assume that for each manipulator $a \in \mathcal{A}$
the grasp transformation matrices $\prescript{g}{}{T}^a_{o} \in \textit{SE(3)}$
from the object's frame to the respective gripper frame are known. We assume that these
grasps are selected such that a stable placement pose can be acquired without releasing the object.

For a pose $\bm{x} \in \textit{SE}(3)$, let $\bm{p_x} = (x, y, z) \in \mathbb{R}^3$ be its position
and $\bm{o_x} = (e_x, e_y, e_z)$ its orientation expressed in rotation angles around the world's $x, y$ and $z$
axis respectively. Our algorithm treats the objective function $\xi(\bm{x})$ as a black-box, however,
we will assume that the function is numerically differentiable w.r.t. the $x, y, e_z$ components of $\bm{x}$.

\section{Method}
\label{sec:method}
We address the problem in \eqref{eq:general_plcmnt} with the framework shown in \figref{fig:overview}.
This framework consists of a pre-processing stage and an optimization stage, \algref{algo:high_level}.
The framework receives the information listed on the left as input and produces
paths $\tau_i: [0, 1] \to \mathcal{C}^a_\text{free}$ for the different arms $a \in \mathcal{A}$ as output.
The final configuration of each path $\tau_i(1)$ represents a placement solution using a particular arm,
and places the object at a stable and collision-free placement pose $\bm{x} = O(\tau_i(1))$. The optimization algorithm, \algref{algo:high_level},
operates in an anytime fashion, meaning that it iteratively produces new solutions $\tau_{i'}$ that
achieve better objective $\xi_i' = \xi(O(\tau_i'(1)))$ than the previous solutions $\tau_i, i< i'$.

The base idea of our approach is to decompose the problem into a search for feasible
placement poses that fulfill all constraints, and only subsequently optimize the objective.
In general, we can not model all of the constraints in \eqref{eq:general_plcmnt}
in closed form. However, for a particular pose $\bm{x} \in \mathcal{X}^o$
we can verify whether it fulfills the constraints. We therefore address the optimization
problem in a sampling-based manner.
For each constraint in \eqref{eq:general_plcmnt} our framework has a component
designed to verify it or to provide samples fulfilling it:

\textbf{Stable placement.} A necessary condition for a pose $\bm{x} \in \mathcal{X}^o$
to be stable, $s(\bm{x})~=~1$, is that the object is in contact with the environment.
Therefore, in the pre-processing stage, we extract surfaces in the target volume and on the object
that afford placing. With these surfaces we can obtain an approximation $\hat{S} \subset \mathcal{X}^o$
of the set of stable placement poses that serves as search space for our optimization.
In addition, these surfaces allow us to verify, whether the object is placed stably
at a given pose.

\textbf{Physically feasible placement.} Within the set $\hat{S}$ we need to locate object poses that
are physically feasible, $c_f(\bm{x}) = 1$, i.e. poses that do not result in penetration of any obstacles.
In addition, we need to verify that these poses can be reached by collision-free arm configurations
$q~\in~\mathcal{C}^a_\text{free}$ for at least one arm, as this
is a necessary condition for a placement pose to be path-reachable, $r(\bm{x})~=~1$.
This verification as well as the sampling is performed within \algref{algo:sample}, which
we refer to as goal sampling algorithm.

\textbf{Reachable placement.} To verify the path-reachability of candidate poses, $r(\bm{x}) =1$, we need to construct approach
paths to them. For this, we employ a sampling-based motion planning algorithm~\cite{Elbanhawi2014}
that receives arm configurations sampled by the goal sampling algorithm as goals.

\textbf{Preferred placement.} The optimization of the objective function is achieved through two concepts.
First, we employ a greedy local optimization algorithm on the poses for which all constraints
have been verified to be fulfilled. Second, whenever the motion planner succeeds in verifying path reachability
for a new pose sample $\bm{x}$, we constrain following iterations to only
validate path-reachability for poses $\bm{x}'$ that achieve a better objective $\xi(\bm{x}') > \xi(\bm{x})$.

\subsection{Defining Potential Contacts}
\begin{figure}[t]
        \includegraphics[width=\columnwidth]{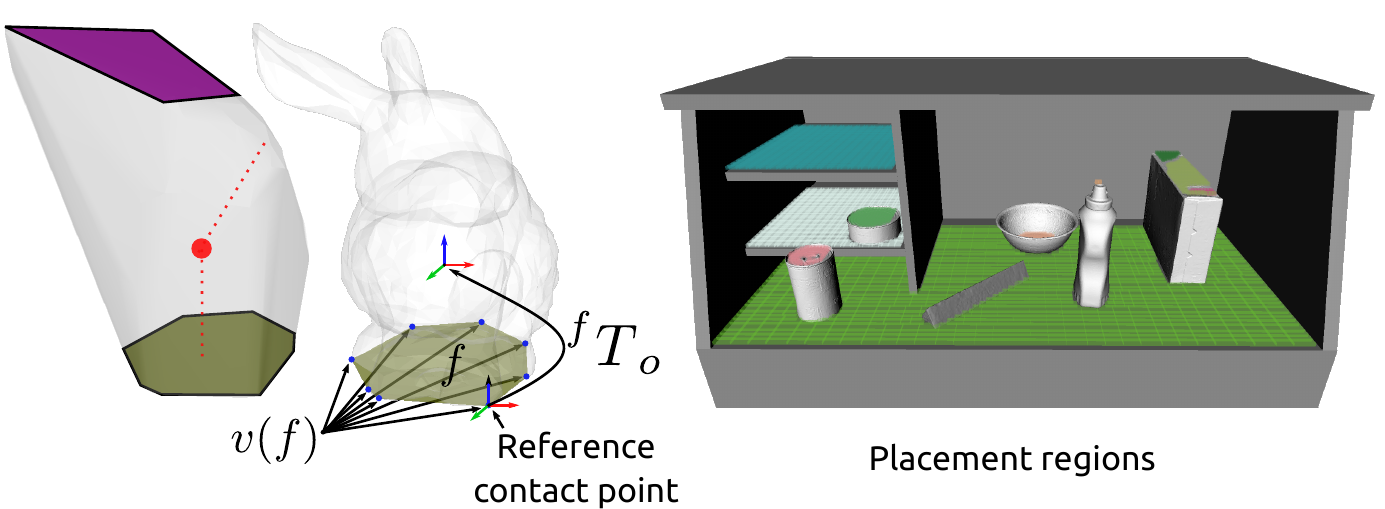}
        \caption{Placement faces and regions. \textit{Left:} The Stanford bunny model is shown with its
                convex hull and two of the hull's faces are highlighted (\textit{purple} and \textit{green}).
                By projecting the center of mass (\textit{red}) along the faces' normals, we
                can determine which face supports a stable horizontal placement. For faces
                $f \in \mathcal{F}$ that support a placement, we refer by $v(f)$ to its vertices and
                by $\prescript{f}{}T_{o}$ to the transformation matrix from a reference vertex to the
                object's frame. \textit{Right:} We extract contiguous horizontal surfaces in the environment
                that provide us with candidate location to place the object at.
                }
        \label{fig:regions_and_faces}
        \vspace{-0.5cm}
\end{figure}

Modeling the set of stable poses $S = \{\bm{x} \in \mathcal{X}^o~|~s(\bm{x}) = 1\}$ of stable placement poses
is challenging, due to the large variety of possible placements.
While we only require samples from this set, rejection sampling on $\mathcal{X}^o$ does not suffice,
due to low probability of satisfying the constraint $s(\bm{x}) =1$. We therefore approximate this set
by the set of poses at which the object is placed on horizontal surfaces.
For this, we extract a discrete set of \textit{placement contact regions},
$\mathcal{R} = \{r_i\}_{i=1}^{i=m}, r_i \subset \mathcal{S}\cap V \subsetneq \mathbb{R}^3$,
from the surface geometry in the target volume.
A placement contact region is a contiguous set of surface points that share the same height,
see \figref{fig:regions_and_faces}. In our implementation, we extract these from an occupancy grid
of the environment, however, also other techniques could be employed.

To determine the orientation in which the object should make contact with these regions,
we extract contact points from the object's surface. For this, we follow a similar approach as in aforementioned
previous works \cite{Wan2018, Lertkultanon2018} and select faces from the object's convex hull to place the
object on.  The convex hull of a finite set of points, i.e.\ a point cloud of
the object, is a convex polyhedron. A face of this polyhedron supports a stable
placement on a horizontal surface, if the projection of the object's center of
mass along the face's normal falls into this face, see
\figref{fig:regions_and_faces}. We refer to the $k \in \mathbb{N}$ number
of faces for which this is the case as
\textit{placement faces}, $\mathcal{F} = \{f_i\}_{i=1}^{i=k}$.

Only the vertices of the boundary of a placement face, $v(f)$, are guaranteed to be part of
the actual object's surface. It is therefore these vertices
that need to be in contact with the support surface in order for the object to be placed stably.
For each face, we select one of these vertices as reference contact point and define a transformation
matrix $\prescript{f}{}T_{o}$ as shown in \figref{fig:regions_and_faces}. With this,
the combination of a contact
region~$r \in \mathcal{R}$ and a placement face~$f \in \mathcal{F}$ defines a class of object poses
\begin{equation*}
        \footnotesize
\setlength{\abovedisplayskip}{4pt}
        \setlength{\belowdisplayskip}{4pt}
        \hat{S}(r, f) =
        \{T\bigl(R_z(\theta), \begin{pmatrix}x \\ y \\ z_r\end{pmatrix}\bigr) \prescript{f}{}T_{o}~|~\begin{pmatrix}x \\ y \\ z_r\end{pmatrix} \in r, \theta \in [0, 2\pi)\},
\end{equation*}
where  $z_r$ is the $z$-coordinate of the placement region, $R_z(\theta)$ the rotation matrix
around the $z$-axis by angle $\theta$, and $T(\cdot, \cdot)$ an operator that combines these
to a transformation matrix. These poses vary in $x, y$ translation within the contact region
and rotation by $\theta$ around the $z$-axis going through the reference point located at $x, y, z_r$.
The union $\hat{S} = \bigcup_{~r \in \mathcal{R}, f \in \mathcal{F}} \hat{S}(r, f)$ of all these
sets is then a parameterized approximation of $S$ that we can search for feasible placements.

Note that not all poses in $\hat{S}$ are stable, since it's definition only guarantees that
the reference contact point is in contact with the placement contact region.
To verify that a pose $\bm{x} \in \hat{S}$ is actually stable, we need to verify that all vertices
of the respective placement face are in contact with a placement region.
They may, however, be in contact with different regions, which allows placements
where, for instance, an object is placed over a gap in the support surface.

\setlength{\textfloatsep}{0pt}
\begin{algorithm}[t]
    \footnotesize
    \SetFuncSty{textsc}
    \DontPrintSemicolon
    \SetKwIF{If}{ElseIf}{Else}{if}{}{else if}{else}{endif}
    \SetKwFor{For}{for}{}{endfor}
    \SetKwFor{While}{while}{}{endw}
    \SetKw{Not}{not}
    \SetKw{Is}{is}
    \SetKw{Publish}{publish}
    \SetKwInOut{Constants}{Constants}
    \SetKwFunction{SampleGoal}{SampleGoals}
    \SetKwFunction{LocalOpt}{OptimizeLocally}
    \SetKwFunction{PlanMotion}{PlanMotion}
    \SetKwFunction{Terminate}{Terminate}

        $M_s, G_s \gets \emptyset, \emptyset$ \tcp*[f]{Storage of internal state} \;
        $\tau, \xi_\text{best}, \mathcal{G} \gets \bot, -\infty, \emptyset$ \;
        \While{\Not \Terminate{}}  {
                $\mathcal{G}_n, G_s \gets$ \SampleGoal{$g_\text{max}, \xi_\text{best}, G_s$} \;
                $\mathcal{G} \gets \mathcal{G} \cup \mathcal{G}_n$ \;
                \If{$|\mathcal{G}| > 0$} {
                        $\tau, M_s \gets$ \PlanMotion{$m_\text{max}, \mathcal{G}, M_s$} \;
                        \If{$\tau \neq \bot$}{
                                $\tau \gets$ \LocalOpt{$\tau$} \;
                                $\tau_\text{best} \gets \tau$ \;
                                $\xi_\text{best} \gets \xi(O(\tau(1)))$ \;
                                $\mathcal{G}_o = \{g \in \mathcal{G} | \xi(O(g)) \leq \xi_\text{best}\}$\;
                                $\mathcal{G} = \mathcal{G} \setminus \mathcal{G}_o$\;
                                \Publish{$\tau_\text{best}, O(\tau_\text{best}(1))$}\;
                        }
                }
        }
        \Return $\tau_\text{best}, O(\tau_\text{best}(1))$ \;
        \caption{High-level Placement Planner}
    \label{algo:high_level}
\end{algorithm}
\subsection{Sampling-based Optimization}
The optimization of the objective is performed by \algref{algo:high_level}.
The algorithm alternates between executing the sub-algorithms \textsc{SampleGoals},
\textsc{PlanMotion} and \textsc{OptimizeLocally} until termination is requested by the user.
\textsc{SampleGoals} computes a finite set of collision-free arm configurations
$\mathcal{G}_n = \{(q, a)~|~ a \in \mathcal{A}, q \in \mathcal{C}^a_\text{free}\}$,
such that for each $q \in \mathcal{G}_n$ the
stability constraint, $s(O(q)) = 1$, and the physical feasibility constraint, $c_f(O(q)) = 1$,
are fulfilled. Furthermore, it only returns configurations for which the placement objective
improves over the best solution found so far, $\xi(O(q)) > \xi_\text{best}$.
Initially, $\xi_\text{best}$ is set to $-\infty$ and is updated,
whenever the motion planner succeeds at finding a new path $\tau$ to
any of the configuration $q \in \mathcal{G}$, where $\mathcal{G}$
stems from the union of all sampled goals.

Motion planning towards the goals, $\mathcal{G}$, is performed in \textsc{PlanMotion}.
The set $\mathcal{G}$ is maintained to only contain configurations
that reach placements of greater objective than $\xi_\text{best}$.
This guarantees that whenever a new path is found, it reaches a placement
with a better objective than any previous solution. Subsequent to finding a new path,
\textsc{OptimizeLocally} is executed to further improve the solution locally.

In each iteration \textsc{SampleFeasible} and \textsc{PlanMotion} receive parameters
$g_\text{max}, G_s, m_\text{max}, M_s$ respectively.
The presence of the parameters $G_s, M_s$ emphasizes that both sub-algorithms
maintain an internal state across iterations of the algorithm,
which is crucial for the efficiency of the overall approach and will be detailed in the following sections.
The parameters $g_\text{max}, m_\text{max}$ limit the computation time budget for each function,
to balance the computational burden of sampling new goals and planning motions to sampled ones.

\subsection{Sampling Kinematically Feasible Placements}
The function \textsc{SampleGoals} needs to solve a constraint satisfaction problem:
\begin{equation}
\begin{aligned}
        & \text{find} & &  a \in \mathcal{A}, q \in \mathcal{C}^a  \\
        & \text{such that}
        & &\xi(O(q)) > \xi_\text{best} \\
        & & &s(O(q)) = 1 \\
        & & &c_f(O(q)) = 1 \\
        & & &q \in \mathcal{C}^a_\text{free}.
\end{aligned}
        \label{eq:csp_kfp}
\end{equation}
This subproblem by itself is challenging to solve. While we only require samples from the
feasible set of this problem, rejection sampling on
$\mathcal{C}^a, a \in \mathcal{A}$ does not suffice, due to low or zero probability of satisfying
$s(O(q)) =1$. Hence, instead we sample the parameterized pose set $\hat{S}$,
for which it likely that $s(\bm{x})$ is fulfilled, and employ an inverse kinematics solver
to compute arm configurations for the sampled poses. This, however, can be rather inefficient in
the presence of obstacles, where the probability
of randomly sampling a collision-free object pose $\bm{x}$ and arm configuration $q \in \mathcal{C}^a$
reaching $\bm{x}$ is low. We remedy this by employing a sampling procedure that adapts its
sampling and focuses on regions of $\hat{S}$ that are likely to fulfill all constraints.

\begin{figure}[t]
        \includegraphics[width=\columnwidth]{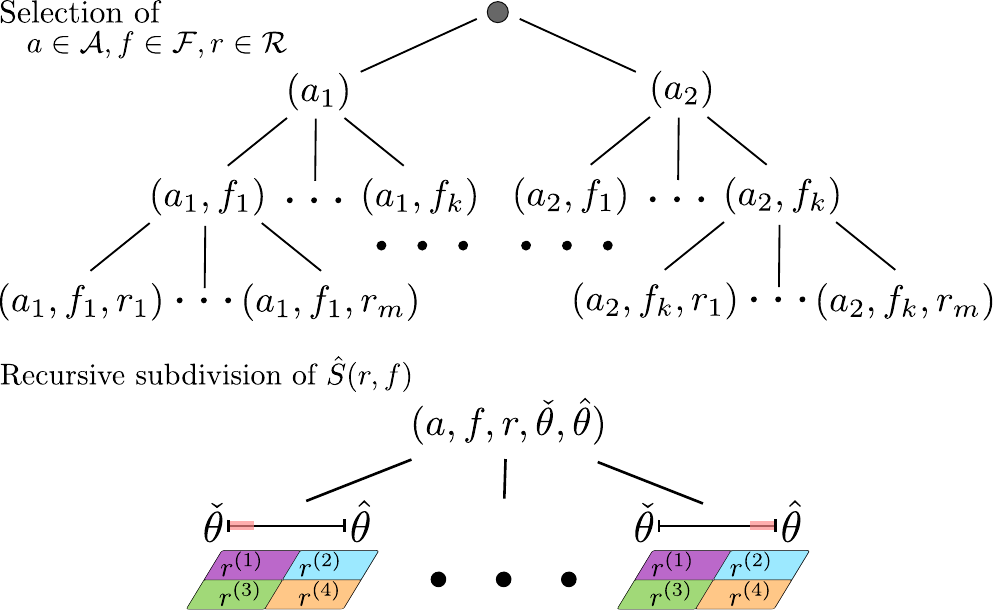}
        \caption{The AFR hierarchy constitutes of two different parts. On the first three level,
                the hierarchy represents choices of an arm $a \in \mathcal{A}$, a placement face
                $f\in \mathcal{F}$ and a region $r \in \mathcal{R}$. On the level at greater depths,
                the hierarchy recursively subdivides the region $r$ and the range of orientations
                $[\check{\theta}, \hat{\theta})$ within a pose set $\hat{S}(r, f)$}
        \label{fig:arf_hierarchy}
\end{figure}

\subsubsection{AFR-Hierarchy}
\label{sec:afr_hierarchy}
Sampling a pose from $\hat{S}$ involves choosing
a placement contact region $r~\in~\mathcal{R}$ and a placement face
$f~\in~\mathcal{F}$. In addition, to compute an arm configuration reaching a
sampled pose, we need to select an arm $a \in \mathcal{A}$. While there is
an overlap of the poses that each arm can reach, some may be more easily
reached by one than the other. Whether a particular placement face $f$ is a
good choice to place an object on depends on the grasp, and thus on
the arm that is selected. Similarly, whether a placement region allows a stable and obstacle penetration
free placement strongly depends on the placement face, as this determines the footprint and
the base orientation of the object.
Furthermore, if a pose $\bm{x} \in \hat{S}(r,f)$ for a particular region $r$ and face $f$
is reachable by an arm $a$, it is likely that poses in close proximity are also reachable by the arm.
Hence, there exists a spatial correlation of feasibility within a set $\hat{S}(r,f)$, as well as between
different sets of $\hat{S}(r,f)$ with similar categorical choices for $r \in \mathcal{R}, f \in \mathcal{F}$
and arms $a \in \mathcal{A}$.

This observation leads us to the definition of the AFR-hierarchy shown in \figref{fig:arf_hierarchy}.
On the first level of this hierarchy, an arm $a \in \mathcal{A}$ is selected, on the second level a
placement face $f \in \mathcal{F}$, and on the third a placement contact region $r \in \mathcal{R}$.
From the third level on, each node in the hierarchy defines all quantities that
we require to sample poses and compute arm configurations. Subsequent level of this hierarchy
recursively partition the sets $\hat{S}(r, f)$. On these lower level, every node represents a
tuple $(a, r, f, \check{\theta}, \hat{\theta})$, where $\check{\theta}, \hat{\theta}$
define a range of orientation angles. The nodes at depth $3$ cover all of
$\hat{S}(r, f)$ and thus it is $\check{\theta} = 0$ and $\hat{\theta} = 2\pi$. The children
of this node, however, will only cover subsets of $\hat{S}(r, f)$ that are constrained in the positions
and orientations.

Let $n^{(i)} = (a, r^{(i)}, f, \check{\theta}^{(i)}, \hat{\theta}^{(i)})$ be a node at depth $i \geq 3$, then
the children of this node arise from subdividing the region $r^{(i)}$ and the interval
$[\check{\theta}, \hat{\theta})$.
The placement region $r^{(i)}$ is divided into four subregions
$r^{(i)} = r^{(i)}_1 \cup r^{(i)}_2 \cup r^{(i)}_3 \cup r^{(i)}_4$ by splitting it along its
mean $x$ and $y$ positions. The interval $[\check{\theta}^{(i)}, \hat{\theta}^{(i)})$ is
split into $l$ equally sized sub-intervals. The resulting $l\times 4$ children
of $n^{(i)}$ then arise from combining each subregion with each interval of the orientation ranges.
This subdivision is continued until some user-specified minimal region area and interval length.

\subsubsection{Monte Carlo Tree Search-based Goal Sampling}
To obtain samples that satisfy \eqref{eq:csp_kfp} we exploit the aforementioned correlation
and employ a Monte Carlo Tree search~(MCTS)~\cite{MCTSSurvey}-based algorithm for sampling.
The algorithm is shown in \algref{algo:sample} and \algref{algo:select}, and uses the
AFR-hierarchy to produce the desired samples. \algref{algo:sample} is the \textsc{SampleGoal}
procedure that is called by \algref{algo:high_level}.

The key idea of the algorithm is that it incrementally constructs a tree of the nodes in
the AFR hierarchy, and stores for each node the proportion of valid samples that
have been obtained from its subbranch. This knowledge is then used
to focus sampling on branches of the hierarchy that are likely to contain valid
samples, while still maintaining some exploration. The tree is stored in the variable $G_s$, and
thus steadily constructed across all executions of \textsc{SampleGoal} in \algref{algo:high_level}.

Every time \algref{algo:sample} is executed it attempts to produce $g_\text{max}$ goal
samples $(q, a)$, where $q \in \mathcal{C}^{a}_\text{free}$ is a collision-free
arm configuration for arm $a$ reaching a feasible placement pose, $s(O(q)) = 1, c_f(O(q)) = 1$
that improves the objective $\xi(O(q)) > \xi_\text{best}$. For each sample, the algorithm first
selects a node $n$ from the AFR hierarchy using \algref{algo:select}. \algref{algo:select} ensures
that this node fully specifies a set $\hat{S}(r, f)$ or a subset thereof as described in
\ref{sec:afr_hierarchy}. It then randomly samples a pose from this set and evaluates whether
the pose constraints $c_f(\bm{x}), s(\bm{x})$ and $\xi(\bm{x}) > \xi_\text{best}$ are fulfilled.
If this is the case, it employs an inverse kinematics solver to compute an arm configuration
$q \in \mathcal{C}^{a_n}$ for the arm $a_n$ specified by the AFR node $n$. If such a configuration
exists and it is collision-free, we obtained a new goal sample that can be provided to the motion
planning algorithm.

\setlength{\textfloatsep}{0pt}
\begin{algorithm}[t]
    \footnotesize
    \SetFuncSty{textsc}
    \DontPrintSemicolon
    \SetKwIF{If}{ElseIf}{Else}{if}{}{else if}{else}{endif}
    \SetKwFor{For}{for}{}{endfor}
    \SetKwFor{While}{while}{}{endw}
    \SetKw{Not}{not}
    \SetKw{Is}{is}
    \SetKwInOut{Symbols}{Symbols}
        \KwIn{Number of maximal iterations $g_\text{max}$,
              best achieved objective value $\xi_\text{best}$,
              state storage $G_s$}
        \KwOut{Feasible placement configurations $\mathcal{G}_n$, state storage $G_s$}
    \SetKwFunction{Select}{SelectAfrNode}
    \SetKwFunction{Sample}{Sample}
    \SetKwFunction{IK}{IKSolver}
    \SetKwFunction{Update}{Update}
        $\mathcal{G}_n \gets \emptyset$\;
        \For{$i \gets 1, \ldots, g_\text{max}$} {
                $n \gets $ \Select{$G_s$}\;
                $\bm{x} \gets $ \Sample{$n$}\;
                \If{$s(\bm{x}) = 1 \wedge c_f(\bm{x}) = 1 \wedge \xi(\bm{x}) > \xi_\text{best}$} {
                        $q \gets $ \IK{$\bm{x}, a_n$}\;
                        \If{$q \in \mathcal{C}^{a_n}_\text{free}$}{
                                $\mathcal{G}_n \gets \mathcal{G}_n \cup \{(q, a_n)\}$\;
                        }
                }
                \Update{$n, G_s, \bm{x}, q, \xi_\text{best}$}\;
        }
        \Return $\mathcal{G}_n, G_s$\;
        \caption{\textsc{SampleGoals}: Monte-Carlo Tree search based sampling algorithm}
    \label{algo:sample}
\end{algorithm}
After each sample step, the tree stored in $G_s$ is updated according to whether we successfully obtained
a new goal sample or not. For each sampled node $n$, we store the following information in $G_s$:
\setlength{\leftmargini}{2.5cm}
\begin{itemize}
        \item $v(n)$, the number of samples obtained from $n$ or any of its descendants
        \item $r(n)$, the sum of all rewards obtained for sampling $n$ or any of its descendants
        \item $\textit{Ch}(n, G_s)$, the children of $n$ that have been added to $G_s$
\end{itemize}
The numbers $v(n)$ and $r(n)$ are updated by the \textsc{Update} function, whereas
$\textit{Ch}(n, G_s)$ is updated within \algref{algo:select}.
When we sampled $n$ we obtain a reward
\begin{equation}
        \Delta r(n) = \begin{cases}
                H(\bm{x}, q, \xi_\text{best}) & \text{if } $n$ \text{ is not a leaf of AFR} \\
                1 & \text{if } v(\bm{x}, q, \xi_\text{best}) = 1 \\
                0 & \text{otherwise}
        \end{cases}
\end{equation}
where $v(\bm{x}, q, \xi_\text{best}) = 1$, if all constraints are fulfilled, i.e.\
$s(\bm{x}) = 1, c_f(\bm{x}) = 1, \xi(\bm{x}) > \xi_\text{best}$ and $q \in \mathcal{C}_\text{free}^a$.
This reward is binary, if $n$ is a leaf of the AFR hierarchy and there are no further
subdivisions of the pose set. If $n$, however, is not a leaf, the reward is
a heuristic value $H(\bm{x}, q, \xi_\text{best}) \in [0, 1]$ that also gives non-zero rewards
to samples that fulfill some, but not all of the constraints.
In any case, the reward is recursively propagated to the ancestors $n'$ of $n$ to update their
respective $v(n'), r(n')$. The number of samples, $v(n')$, is always increased by one
as we acquired a single sample, whereas the accumulated reward $r(n')$ is updated by the
reward $\Delta r(n)$ obtained by the sampled node $n$.

The decision on which node to sample is made in the \textsc{SelectAFRNode} function, \algref{algo:select}.
The algorithm always starts at the root of the AFR hierarchy
and descends to a node in the hierarchy that it decides to sample.
Since we can produce samples only for nodes at depths greater than $3$, the algorithm
always needs to descend at least to depth $3$ before returning any node. For nodes $n$ at depth greater
than $3$ the algorithm descends to its children, as long as $n$ is not a leaf
and $n$ has been sampled before.

Initially, $G_s$ only contains the root of the AFR hierarchy. Hence,
the only option the algorithm has is to select a child in the AFR hierarchy that
is not in $G_s$ yet. This is done by the \textsc{AddChild} operation, which selects a random
child from the AFR hierarchy and adds it to $G_s$. Let $n$ now
be any selected AFR node that is already stored in $G_s$. We distinguish between its children $\textit{Ch}(n, G_s)$
that are also stored in $G_s$ and its total set of children $Ch(n)$ as defined by the hierarchy. If $n$ has children
in $G_s$, the algorithm may either descend to one of them, or add a new child to $G_s$, if $|\textit{Ch}(n, G_s)| < |Ch(n)|$.
The decision on what to do is based on the UCB1 policy~\cite{Auer2002}, which is common to
employ in Monte-Carlo Tree search and shown in \algref{algo:select}. It allows to the algorithm to
balance between re-sampling branches (exploitation) that have led to valid samples before
and exploring new branches. The score $u'$ for adding a new child is based on the conservative
assumption that any unsampled node is as good as the average of its siblings.

\begin{figure*}[ht]
        \setlength{\tabcolsep}{2pt}
        \center
        \begin{tabular}{ccc}
                \begin{subfigure}[b]{0.32\textwidth}
                        \includegraphics[width=\textwidth]{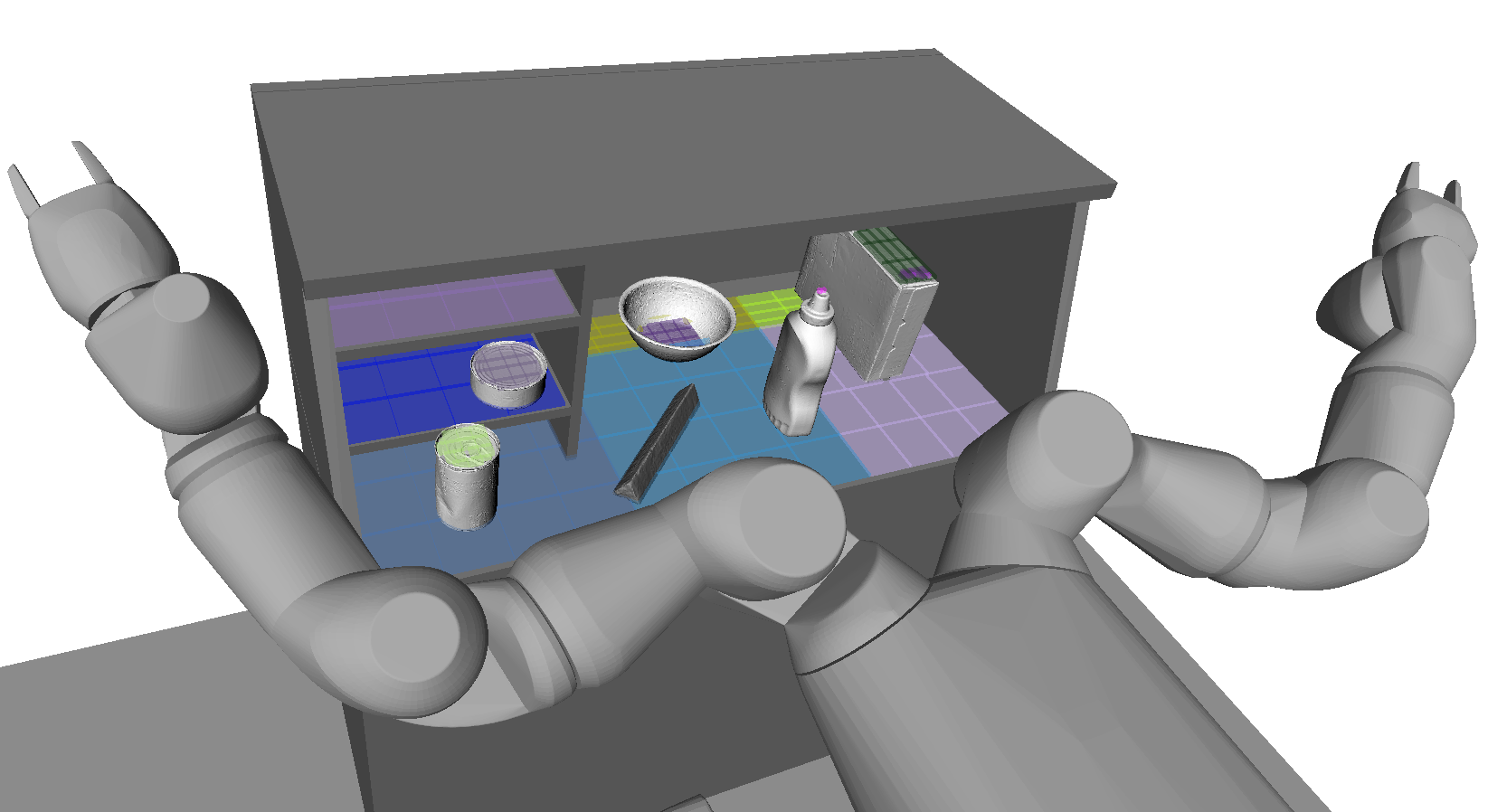}
                   \caption{Scene 1}
                   \label{fig:results:scene_1}
                \end{subfigure}
                &
                \begin{subfigure}[b]{0.32\textwidth}
                        \includegraphics[width=\textwidth]{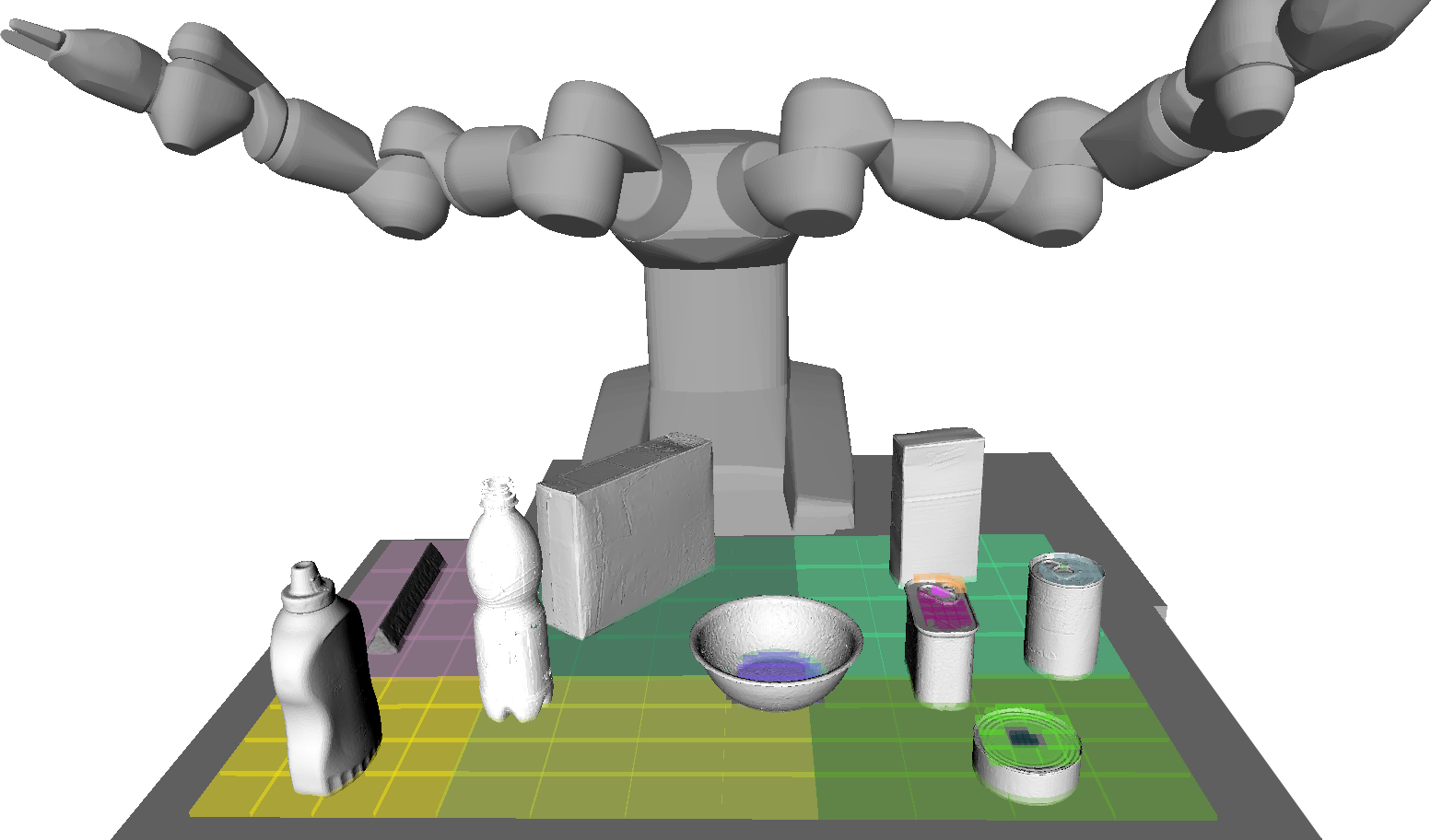}
                        \caption{Scene 2}
                        \label{fig:results:scene_2}
                \end{subfigure}
                &
                \begin{subfigure}[b]{0.32\textwidth}
                        \center
                        \includegraphics[width=\textwidth]{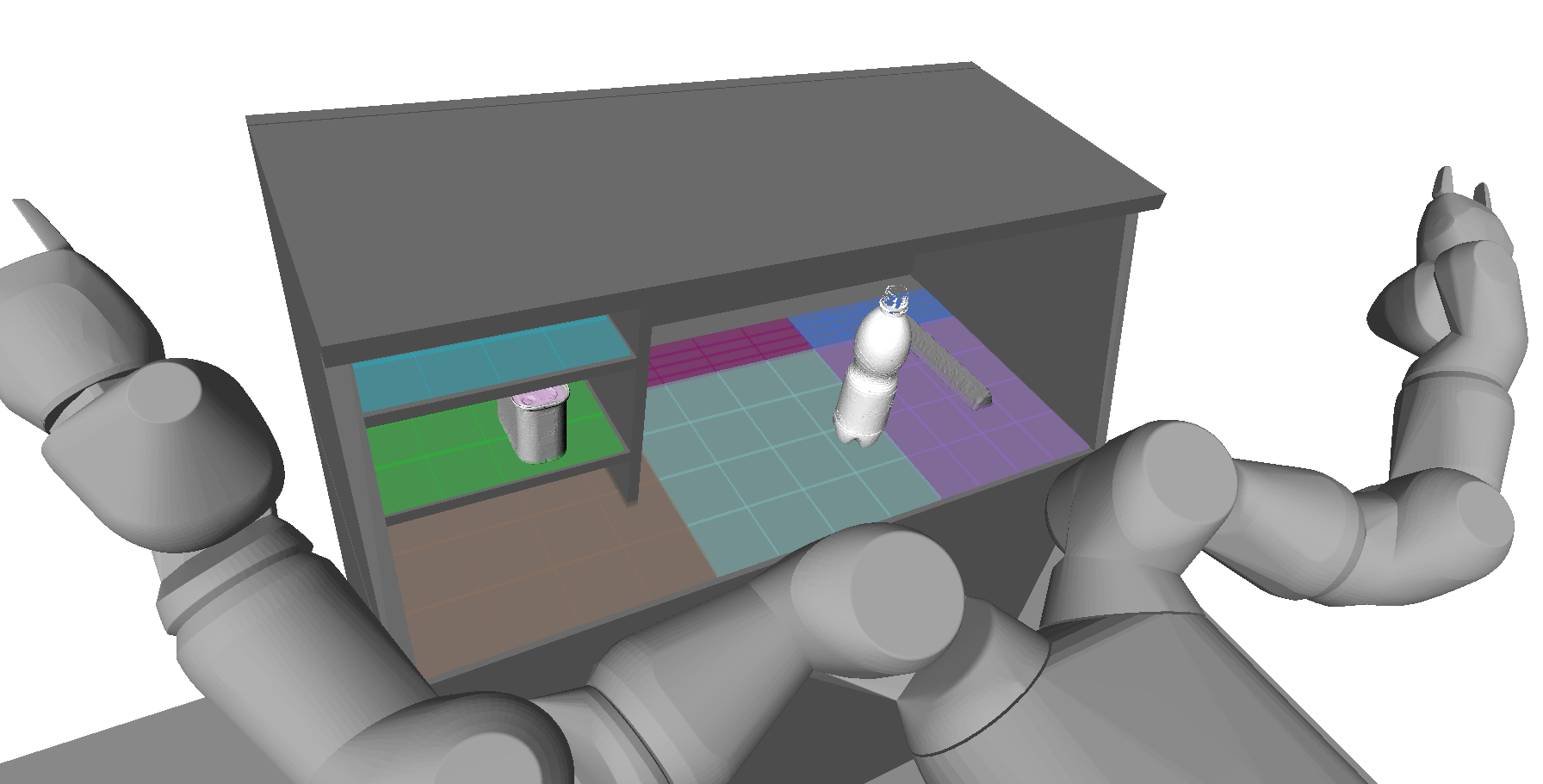}
                   \caption{Scene 3}
                   \label{fig:results:scene_3}
                \end{subfigure}
                \\
                \begin{subfigure}[b]{0.32\textwidth}
                        \includegraphics[width=\textwidth]{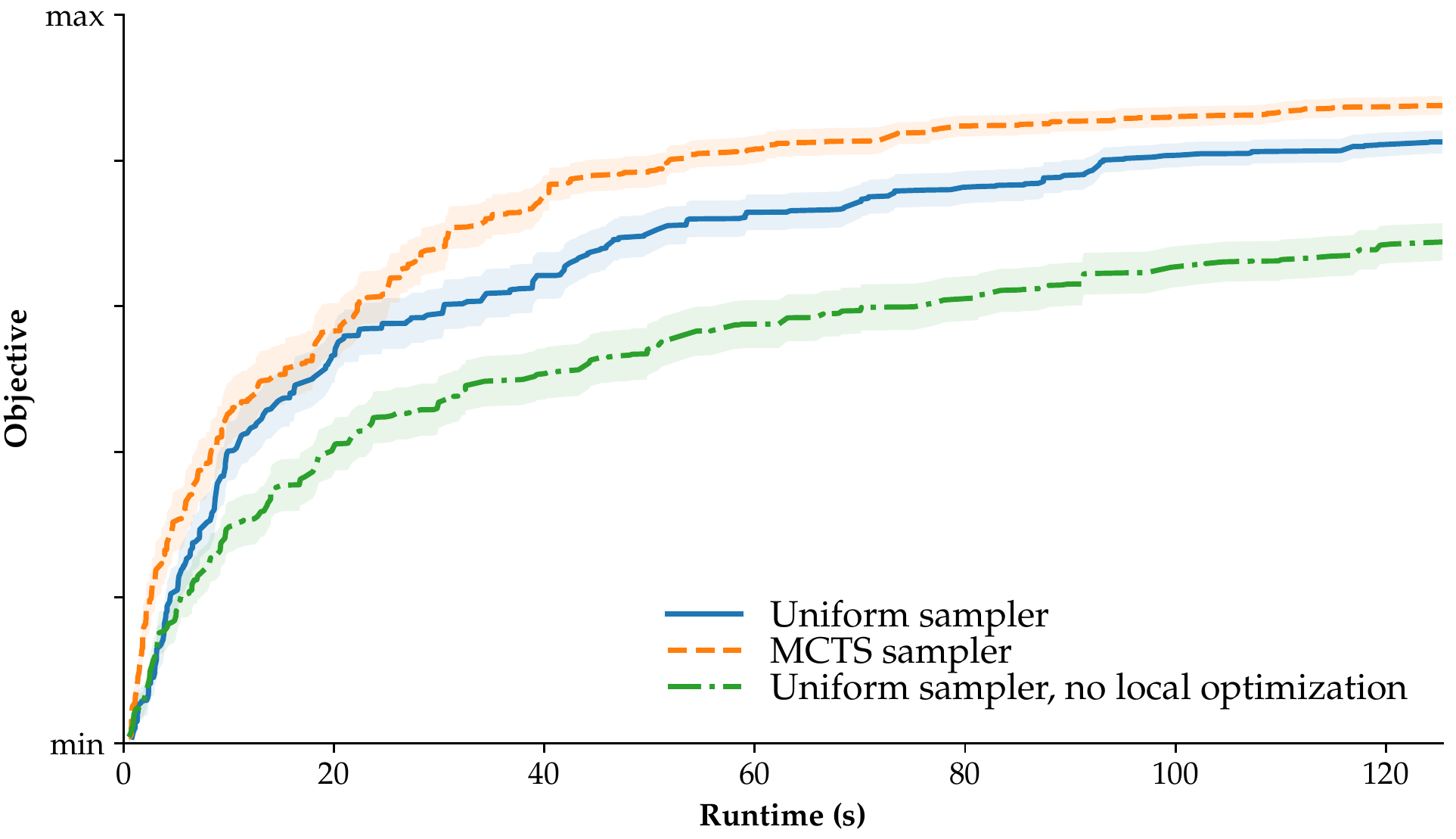}
                        \caption{Average optimization performance for all objects when \textbf{minimizing} clearance in scene 1.}
                   \label{fig:results:scene_1}
                \end{subfigure}
                &
                \begin{subfigure}[b]{0.32\textwidth}
                        \includegraphics[width=\textwidth]{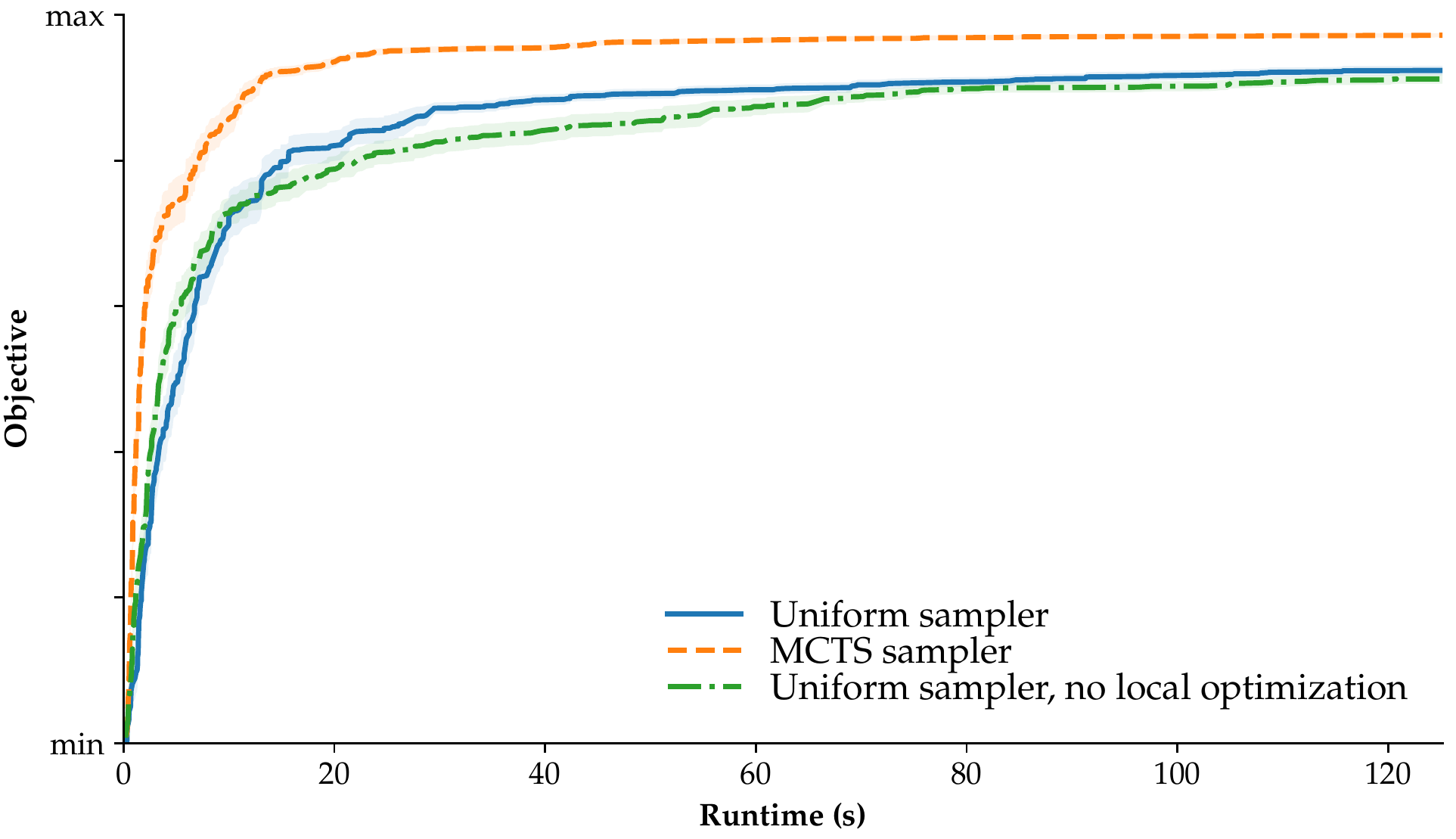}
                        \caption{Average optimization performance for all objects when \textbf{minimizing} clearance in scene 2.}
                        \label{fig:results:scene_2}
                \end{subfigure}
                &
                \begin{subfigure}[b]{0.32\textwidth}
                        \center
                        \includegraphics[width=\textwidth]{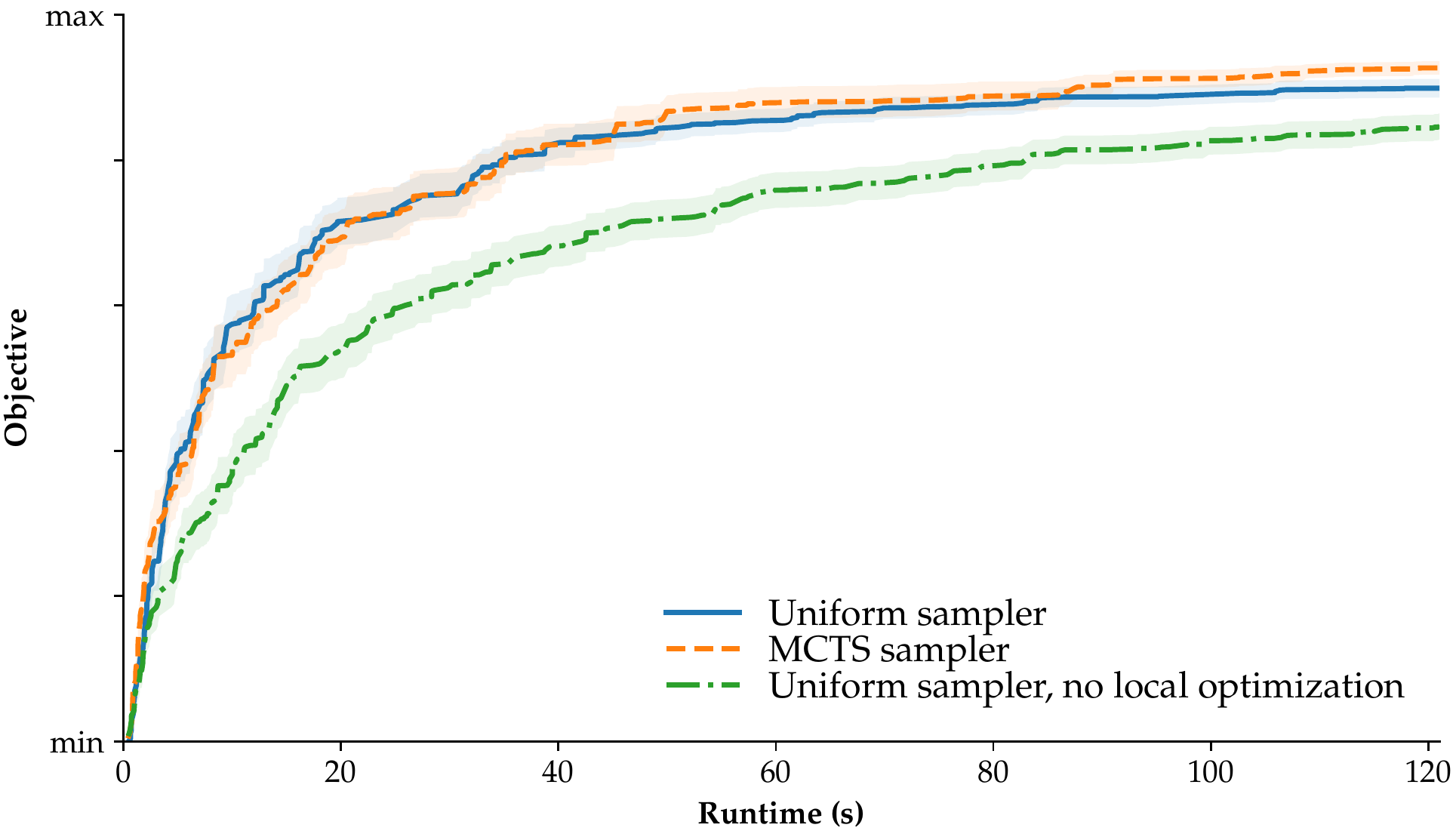}
                        \caption{Average optimization performance for all objects when \textbf{minimizing} clearance in scene 3.}
                   \label{fig:results:scene_3}
                \end{subfigure}
                \\
                \begin{subfigure}[b]{0.32\textwidth}
                        \includegraphics[width=\textwidth]{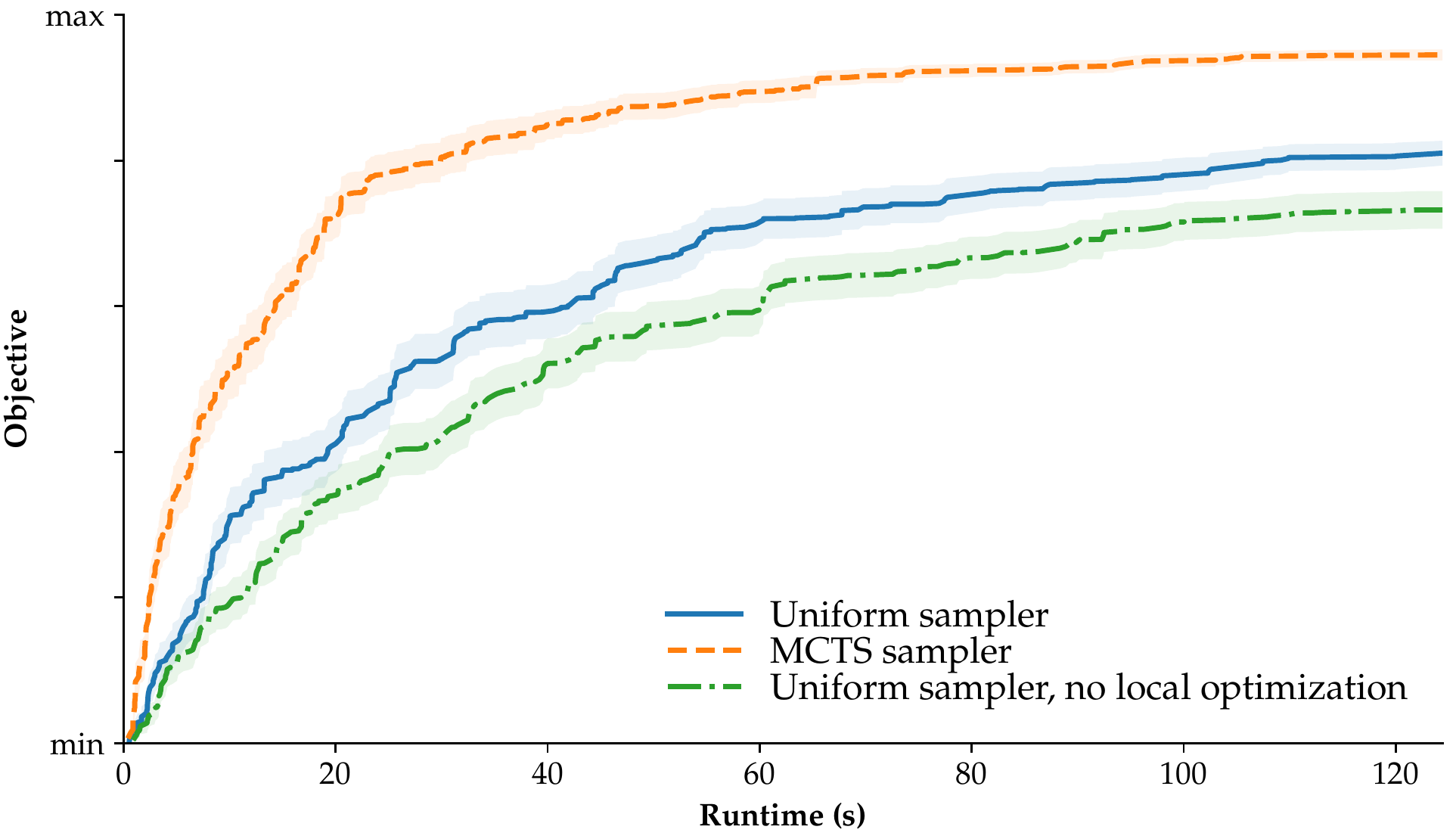}
                        \caption{Average optimization performance for all objects when \textbf{maximizing} clearance in scene 1.}
                   \label{fig:results:scene_1}
                \end{subfigure}
                &
                \begin{subfigure}[b]{0.32\textwidth}
                        \includegraphics[width=\textwidth]{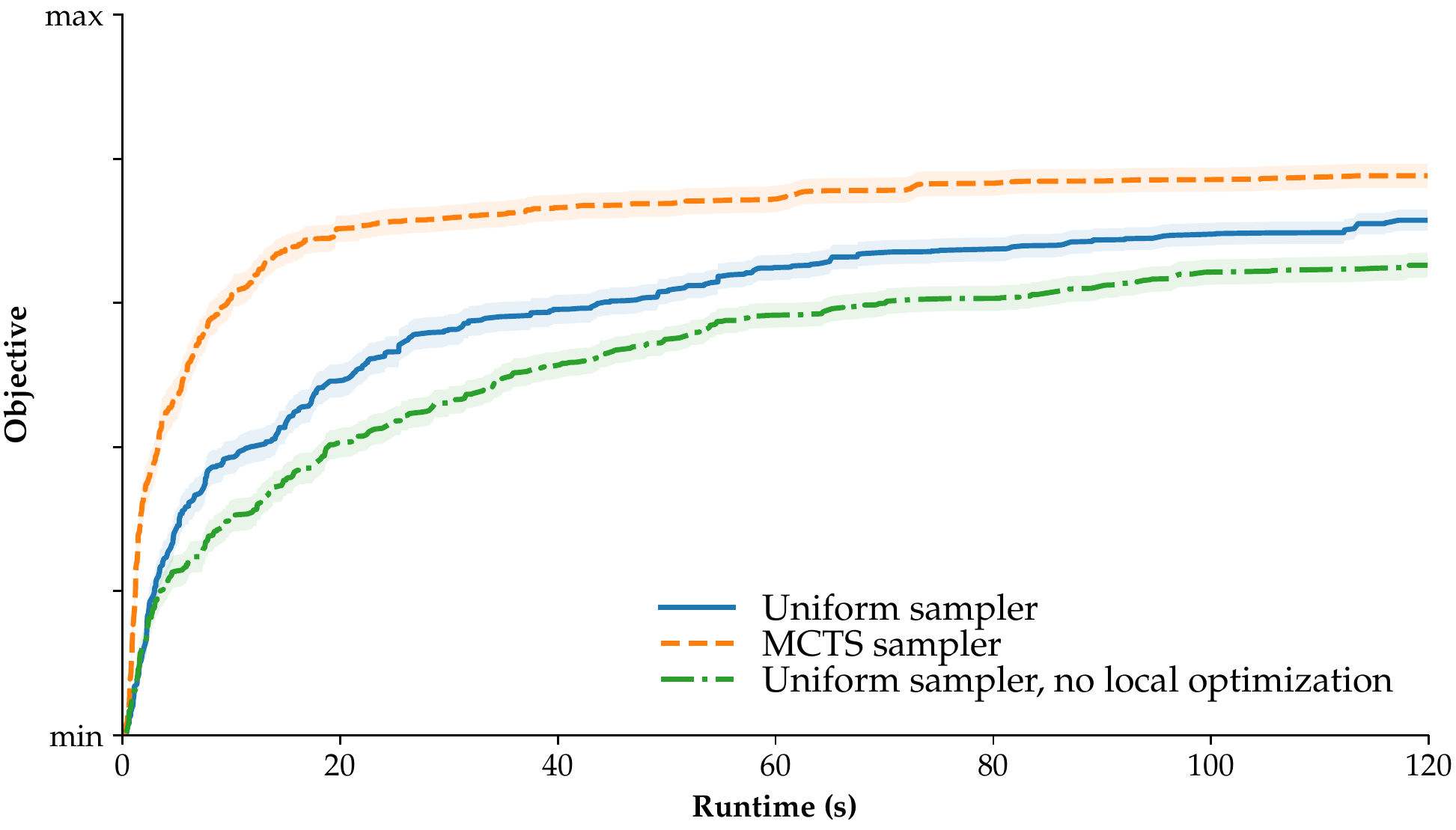}
                        \caption{Average optimization performance for all objects when \textbf{maximizing} clearance in scene 2.}
                        \label{fig:results:scene_2}
                \end{subfigure}
                &
                \begin{subfigure}[b]{0.32\textwidth}
                        \center
                        \includegraphics[width=\textwidth]{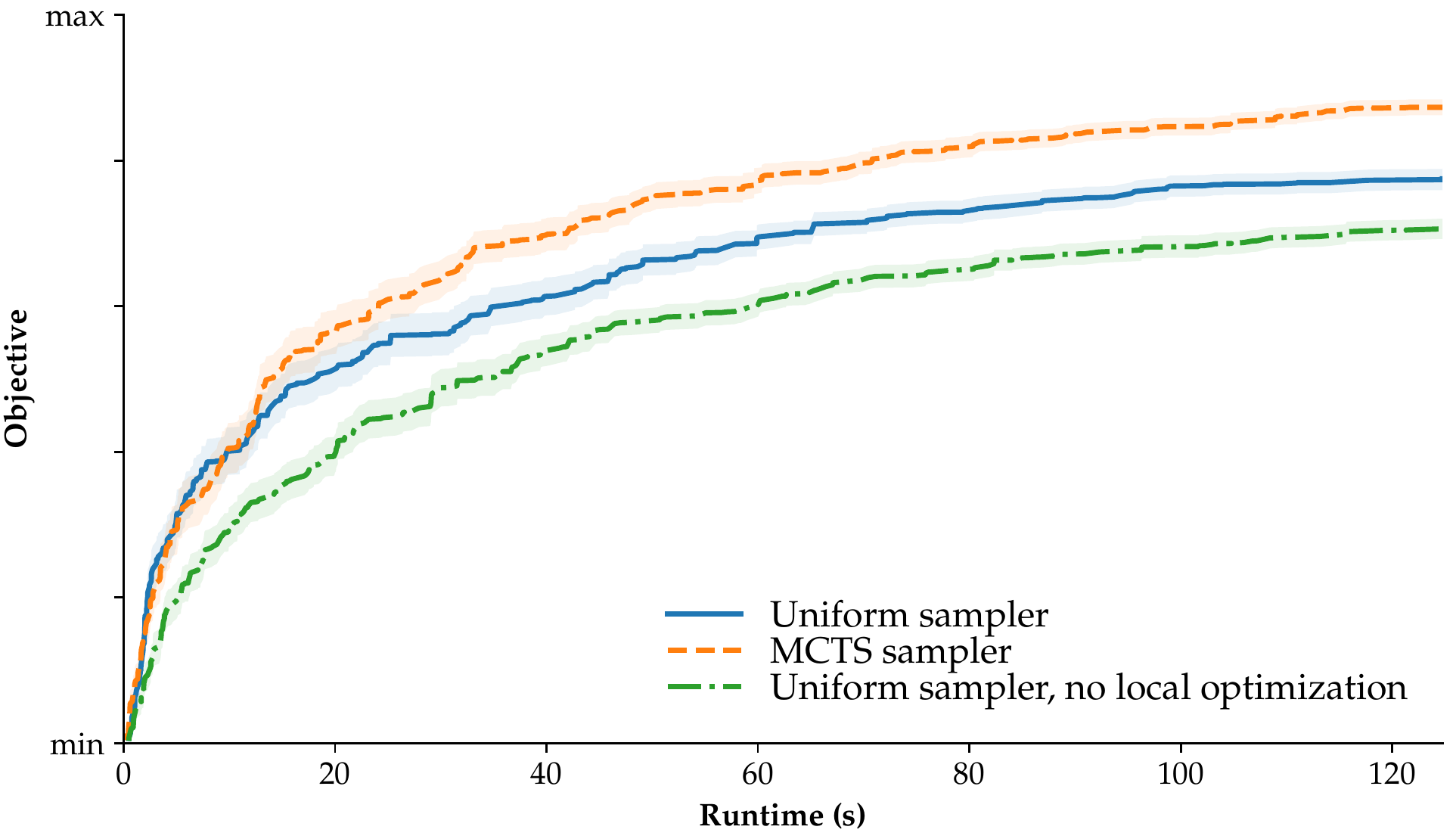}
                        \caption{Average optimization performance for all objects when \textbf{maximizing} clearance in scene 3.}
                   \label{fig:results:scene_3}
                \end{subfigure}
                \\
        \end{tabular}
        \caption{
                Experiments scenes and optimization performance of different instances of our algorithm.
                The plots in \textbf{(d)} - \textbf{(i)} show the mean relative objective achieved by
                each algorithm as a function of planning time. The plots show the average optimization
                performance across all test objects. In order to make the objective values comparable,
                we normalize the achieved objective values for each scene and object into the range
                of minimal and maximal objective observed throughout all executions.
        }
        \label{fig:scenes_plots}
        \vspace{-0.5cm}
\end{figure*}

\setlength{\textfloatsep}{0pt}
\begin{algorithm}[t]
    \footnotesize
    \SetFuncSty{textsc}
    \DontPrintSemicolon
    \SetKw{Not}{not}
    \SetKw{Is}{is}
    \SetKwIF{If}{ElseIf}{Else}{if}{}{else if}{else}{endif}
    \SetKwFor{For}{for}{}{endfor}
    \SetKwFor{While}{while}{}{endw}
    \SetKwInOut{Symbols}{Symbols}
        \KwIn{State storage $G_s$}
        \KwOut{node $n$ in $G_s$ that defines a tuple $(a, f, r, \check{\theta}, \hat{\theta})$}
    \SetKwFunction{Root}{Root}
    \SetKwFunction{SelectChild}{SelectChild}
    \SetKwFunction{IsLeaf}{IsLeaf}
    \SetKwFunction{AddChild}{AddChild}
    \SetKwProg{function}{Function}{}{}
        \function{\SelectChild{$n, G_s$}} {
                \For{$i \in \textit{Ch}(n, G_s)$}{
                    $u_i \gets \frac{r(i)}{v(i)} + c \sqrt{\frac{2\ln(v(n))}{v(i)}}$\;
                }
                $u' \gets -\infty$\;
                \If{$|\textit{Ch}(n, G_s)| < |\textit{Ch}(n)|$} {
                        $u' \gets \frac{1}{j}\sum_{i = 1}^{j}\frac{r(i)}{v(i)} + c\sqrt{\frac{2\ln(v(n))}{j}}$\;
                }
                \If{$\forall i \in \textit{Ch}(n, G_s): u' > u_i \vee |\textit{Ch}(n, G_s)| = 0$}{
                        \Return \AddChild{$n, G_s$}\;
                }
                \Return $\wtfargmax{i \in \textit{Ch}(n, G_s)} u_i$\;
        }

        $n \gets$ \Root{$G_s$}\;
        \For{$d \gets 1 \ldots 3$} {
             $n \gets$ \SelectChild{$n, G_s$}\;
        }
        \While{$v(n) > 0 \wedge \neg$\IsLeaf{$n$}} {
                $n \gets $ \SelectChild{$n, G_s$}\;
        }
        \Return $n$ \;
        \caption{\textsc{SelectAfrNode}: Selection of node AFR-hierarchy}
    \label{algo:select}
\end{algorithm}

\subsection{Motion Planning}
The subalgorithm \textsc{PlanMotion} plans motions for each arm separately,
as we assume that only one arm is required to perform the actual placement,
while all other arms remain in a resting position.
In principle, any motion planning algorithm could be employed for this subalgorithm.
The only requirement on the algorithm is the possibility to efficiently add and remove goal
configurations from the goal set $\mathcal{G}$, desirably without loosing information, e.g. samples
in a search tree, that could be beneficial for planning paths to future goals.

In our implementation, we employ a modification of OMPL's~\cite{Ompl} bidirectional
RRT algorithm~\cite{Kuffner2000}. The algorithm constructs a single forward tree and one
backward tree for each goal in $\mathcal{G}$. Whenever the algorithm succeeds in connecting
the forward tree with a backward tree, the two trees are merged and success is reported. When
$\mathcal{G}$ is modified, the backward trees rooting in goal configurations that have been removed
are still maintained, as they may still prove valuable to reach other goals. When connecting
to any of these, however, the algorithm no longer reports success and only merges it into
the forward tree.

\subsection{Local Optimization}
Whenever the motion planning algorithm succeeds in computing a new path $\tau_i$, we locally
optimize the reached placement pose by following the gradient of the objective function $\xi$.
For this, let $q = \tau_i(1)$ be the final configuration of the path that reaches a placement
pose $O(q) = \bm{x}$. We can locally improve the solution using the following update rule:
\begin{align*}
        \Delta q& \gets J^\dagger v(\frac{\partial \xi}{\partial x, y, e_z}(O(q)))\\
        q& \gets q + \mu\Delta q,
\end{align*}
where $J^\dagger$ is the pseudo-inverse of the arm's Jacobian at $q$, $\mu \in \mathbb{R}^{>0}$
a step size, and $v(x, y, \theta) = (x, y, 0, 0, 0, \theta)^T$ lifts the three dimensional gradient to
a six dimensional end-effector velocity. As long as the updated $q$ is collision-free and $O(q)$ is
not violating any constraints, we concatenate the new configurations to the path $\tau$ and obtain
an improved feasible solution.

    \section{Experiments}
\label{sec:experiments}
We implemented our approach in Python using OpenRAVE~\cite{diankov_thesis}
and the Open Motion Planning Library~\cite{Ompl}.
For evaluation, we plan and optimize placements for four different objects on three different
environments with varying degree of clutter, see \figref{fig:figure_one} and \figref{fig:scenes_plots}.
The objects differ in size, shape and in number of placement faces.
As robot model, we employ ABB's dual-arm robot Yumi, where each arm has $7$ DoFs.
All experiments were run on an Intel Core i7-4790K CPU @ 4.00GHz$\times4$ with $16$GB RAM running
Ubuntu 18.04.

As objective function we employ two variations of clearance to obstacles within the placement volume.
We define the clearance to obstacles as:
\begin{equation}
        C(\bm{x}) = \frac{1}{|\mathcal{B}_o(\bm{x})|} \sum_{\bm{p}' \in \mathcal{B}_o(\bm{x})} d_{\mathcal{S}}(\bm{p'}),
\end{equation}
where $\mathcal{B}_o(\bm{x})$ denotes a finite set of points approximating the volume
of $o$ when it is located at pose $\bm{x}$.
The function $d_{\mathcal{S}}: \mathbb{R}^3 \to \mathbb{R}$
denotes the distance in $x,y$ and positive $z$ direction to the environment's surface
within the target volume $V$.
Maximizing this function, i.e.\ $\xi(\bm{x}) = C(\bm{x})$, leads to placements where
the object is distant to obstacles. This is useful, for example, if the robot is tasked
with manipulating the object further after placing. Minimizing this clearance function,
i.e.\ $\xi(\bm{x}) = -C(\bm{x})$, on the other hand, is a good heuristic when the robot
is tasked to pack multiple objects into a limited volume. This objective is particularly interesting, as
the closer the object is to be placed to obstacles, the more difficult it is to obtain
a collision-free approach path.

As can be seen in \figref{fig:figure_one}, and better so in the accompanying video, our algorithm
succeeds at computing placements with high objective values for all test cases. To evaluate the
planning and optimization performance of our approach, we compare it to two simplified modifications.
In the first modification, we replace
the Monte-Carlo Tree search-based sampling algorithm, \algref{algo:sample}, with a naive uniform sampler.
In addition, in the second modification we remove the local optimization from \algref{algo:high_level}.

We ran each instance for each objective, object and scene $20$ times for $2$ minutes
and recorded the objective values of the found solutions. The progress of the
average objective values as a function of runtime is shown in \figref{fig:scenes_plots}.

In all test cases all instances of our algorithm compute initial solutions within a few seconds,
and succeed at locating better solutions as time progresses. We observe that our
algorithm using MCTS performs better than, or as good, as the baselines. The uniform
sampler without local optimization performs worse than the one with. Hence, we conclude
it is both the local optimization and the MCTS sampling that enable our algorithm to find better solutions
faster, and on average achieve larger final objectives.

The mean objective values increase quickly in the beginning
before slowing down as they approach the maximum objective ever observed in the respective scene.
This is likely due to the fact that the probability of locating poses that improve
the objective declines the higher the best achieved objective is.

    \section{Discussion $\And$ Conclusion}
\label{sec:discussion}
We presented an algorithmic framework that computes robot motions
to transport a grasped object to a stable placement that optimizes a user-provided
objective. Our approach is capable of achieving this
even in environments cluttered with obstacles. The approach considers all
available arms to reach the best placement and operates in an anytime-fashion,
computing initial low-objective solutions quickly and improving on it as computational
resources allow.

Our approach combines sampling-based motion planning and local optimization
with a hierarchical sampling algorithm based on MCTS. A key novelty
lies in the AFR hierarchy and applying MCTS as sampling algorithm. While our
experiments already demonstrate the advantage of this approach, we believe
it has more potential. For instance, in this work the grasp, that we place the object
with, is determined by the chosen arm. An interesting future extension is to incorporate
different choices of grasps into the hierarchy. In addition, the sampling algorithm
could employ pruning of branches of the hierarchy, if a branch can be proven to
not contain any solutions.

Further, we believe the approach can be extended to different types of placements.
The different combinations of placement faces and regions constitute disjoint contact classes
of object poses. In future work, we intend to extend the AFR hierarchy to more diverse
contact classes, such as an object leaning against a wall.

Lastly, one of the weaknesses of the sampling-based optimization approach is the decreasing
convergence rate observed in our experiments. To remedy this, we intend to investigate whether
we can exploit gradient information not only in the local optimization step, but also in the
goal sampling algorithm.

\vspace{-0.415cm}

    \bibliographystyle{ieee/IEEEtran}
    \bibliography{ieee/IEEEabrv,placement}
\end{document}